\documentclass[conference]{IEEEtran}
\IEEEoverridecommandlockouts

\usepackage{cite}

\ifCLASSINFOpdf
\else
\fi
\usepackage{amsmath}
\usepackage{amssymb}  
\usepackage{amsfonts}
\usepackage{dsfont}
\usepackage{mathtools}
\usepackage{nicefrac}
\mathtoolsset{showonlyrefs}
\usepackage{float}
\ifCLASSOPTIONcompsoc
  \usepackage[caption=false,font=normalsize,labelfont=sf,textfont=sf]{subfig}
\else
  \usepackage[caption=false,font=footnotesize]{subfig}
\fi
\usepackage{url}

\usepackage[german,english]{babel}
\usepackage{balance}

\include{math}

\def\blind#1{#1}

\usepackage{dblfloatfix}

\usepackage[nohyperlinks, printonlyused]{acronym}
\usepackage{dsfont}
\usepackage{array}

\newcolumntype{P}[1]{>{\centering\arraybackslash}p{#1}}
\newcolumntype{R}[1]{>{\raggedleft\arraybackslash}p{#1}}
\newcolumntype{L}[1]{>{\raggedright\arraybackslash}p{#1}}

\usepackage{multirow}
\usepackage{relsize}

\usepackage{makecell}

\usepackage{amsmath}
\usepackage{svg}
\usepackage{adjustbox}
\usepackage{acronym}

\usepackage{color, colortbl}

\definecolor{custom-grey}{HTML}{c7c7c7}
\definecolor{custom-grey2}{HTML}{a1bcd4}

\definecolor{light-grey}{HTML}{e6e6e6}
\definecolor{greena5}{HTML}{28bf32} 
\definecolor{greena1}{HTML}{a2c48b} 

\definecolor{green1}{HTML}{fff05a}
\definecolor{green2}{HTML}{ffd25a}
\definecolor{green3}{HTML}{ffaa5a}
\definecolor{green4}{HTML}{ff785a}
\definecolor{green5}{HTML}{e76244} 

\definecolor{red3}{HTML}{ce501f}
\definecolor{red2}{HTML}{ce7856}
\definecolor{red1}{HTML}{ce8f76} 

\newacro{2D}[2D]{two-dimensional}
\newacro{DT}[DT]{decision tree}
\newacro{EM}[EM]{Expectation-Maximization}
\newacro{FG}[FG]{full gating}
\newacro{IML}[IML]{interpretable machine learning}
\newacro{ML}[ML]{machine learning}
\newacro{MoDT}[MoDT]{Mixture of Decision Trees}
\newacro{MoE}[MoE]{Mixture of Experts}
\newacro{RF}[RF]{random forest}
\newacro{RSS}[RSS]{residual sum of squares}

\usepackage{xargs}
\usepackage[colorinlistoftodos,prependcaption]{todonotes}
\newcommandx{\pascal}[1]{\todo[linecolor=blue,backgroundcolor=blue!25,bordercolor=blue,inline,caption={P}]{#1}}
\newcommandx{\pk}[1]{\todo[linecolor=blue,backgroundcolor=blue!25,bordercolor=blue,caption={P},size=\scriptsize]{#1}}

\begin{document}
%
\title{Mixture of Decision Trees for\\Interpretable Machine Learning}


\blind{
\author{\IEEEauthorblockN{Simeon Br\"uggenj\"urgen and Pascal Kerschke}
\IEEEauthorblockA{
Department of Information Systems,\\
University of Münster, Germany\\
Institute of Transport and Economics,\\
Dresden University of Technology\\
Email: {\tt\small simeon.brueggenjuergen@uni-muenster.de}}
\and
\IEEEauthorblockN{Nina Schaaf and Marco F. Huber}
\IEEEauthorblockA{Department Cyber Cognitive Intelligence (CCI), Fraunhofer IPA \\Stuttgart, Germany\\
Institute of Industrial Manufacturing and Management~IFF\\
University of Stuttgart, Germany\\
Email: {\tt\small marco.huber@ieee.org}}
}
\author{Simeon Br\"uggenj\"urgen$^{1}$, Nina Schaaf$^{2}$, Pascal Kerschke$^{3}$, and Marco F. Huber$^{2,4}$
\thanks{$^{1}$Simeon Br\"uggenj\"urgen is with the Department of Information Systems, University of M\"unster, Germany.\newline
{\tt\small simeon.brueggenjuergen@uni-muenster.de}}%
\thanks{$^{2}$Nina Schaaf and Marco F. Huber are with the Department Cyber Cognitive Intelligence (CCI), Fraunhofer Institute for Manufacturing Engineering and Automation IPA, Stuttgart, Germany.
{\tt\small marco.huber@ieee.org}}%
\thanks{$^{3}$Pascal Kerschke is with the Institute of Transport and Economics, Dresden University of Technology, Germany.
}%
\thanks{$^{4}$Marco F. Huber is with the Institute of Industrial Manufacturing and Management IFF, University of Stuttgart, Germany.\newline
}%
}
}


%


\maketitle


\begin{abstract}


This work introduces a novel interpretable machine learning method called \ac{MoDT}. It constitutes a special case of the Mixture of Experts ensemble architecture, which utilizes a linear model as gating function and decision trees as experts. Our proposed method is ideally suited for problems that cannot be satisfactorily learned by a single decision tree, but which can alternatively be divided into subproblems. Each subproblem can then be learned well from a single decision tree. Therefore, \ac{MoDT} can be considered as a method that improves performance while maintaining interpretability by making each of its decisions understandable and traceable to humans.

Our work is accompanied by a Python implementation, which uses an interpretable gating function, a fast learning algorithm, and a direct interface to fine-tuned interpretable visualization methods. The experiments confirm that the implementation works and, more importantly, show the superiority of our approach compared to single decision trees  and random forests of similar complexity.

\end{abstract}


%
\section{Introduction}
\label{sec:introduction}

\Ac{ML} has been and continues to be an enormous success story. Historically, it has been the goal of researchers to develop problem-solving ML algorithms with ever-increasing accuracy. 
However, in recent years, the public and research interest in \ac{IML} and its overarching field explainable artificial intelligence (XAI) has soared \cite{Molnar2019, samek2019explainable}.
This development is fueled by the desire of humans to understand how algorithms and machines operate in an increasingly digitized and data-driven world. While algorithms are developed that continuously claim more domains where they are outperforming humans, the inner workings of these algorithms tend to become more complex. As humans are arguably not even able to focus on more than one task at once, it becomes clear that it is almost impossible to comprehensively understand the plethora of internal factors that contribute to a complex model’s decision. Still, humans want to use, evaluate, and trust algorithms. To accommodate the two latter desires, methods of \ac{IML} can be used to simplify, distill, and translate algorithmic decisions into a human-comprehensible format.

There is a plethora of methods in the field of \ac{IML} that can be categorized along multiple dimensions. A typical criterion is whether an \ac{IML} method can be used on any underlying \ac{ML} model (model-agnostic), or if it is only suitable for a specific type of model (model-specific) \cite{Carvalho2019}. On another dimension, IML methods that are used after the creation of the to-be analyzed ML model are grouped under the term post-hoc interpretability \cite{Laugel2019,Lipton2018,Carvalho2019}. Examples of model-agnostic post-hoc methods are LIME \cite{lime} and SHAP \cite{shap}.
In contrast, there are \ac{IML} methods where interpretability is already gained at the time of the creation of the ML model. These intrinsically interpretable models are by definition always model-specific \cite{Molnar2019, JAIR2021}. Examples of an intrinsically interpretable method are a (simple) linear regression, \acp{DT}, or the here presented new method. 

A \ac{DT} is an ML technique that can be highly interpretable. This is due to the simple and intrinsic way to visualize a \ac{DT} model and the ability to decompose it into easily understandable decision rules. A \ac{DT} can be used to solve ML problems directly, however, it is not the most powerful method. Depending on the problem, there can be alternatives like deep neural networks or random forests that achieve higher performance but are in turn not interpretable anymore~\cite{JAIR2021}. 
Therefore, research aims at developing algorithms that exceed the performance limitations of \acp{DT} but still remain interpretable.
An approach by Vasic et al. \cite{Vasic2019} successfully merges DTs and the \ac{MoE} architecture. \ac{MoE} is an ensemble technique that was initially proposed in 1991 by Jacobs et al. \cite{Jacobs1991}. It uses a so-called gating function to assign the input data to distinct experts. While there have been plenty of publications that explore the possibilities of \ac{MoE}, the technique remains a niche. Vasic et al. \cite{Vasic2019} demonstrate that it is possible to successfully combine the technique with \acp{DT}, however, do not discuss the implications for \ac{IML}.


In this paper we introduce a variant of the MoE architecture using \acp{DT} that specifically focuses on interpretability. 
To the best of our knowledge, such an approach has not been published before. A ready-to-be-used implementation of this approach is written in Python and can be found on GitHub\footnote{\url{https://github.com/simsal0r/mixture-of-decision-trees}}. In order to highlight the methods' reliance on \acp{DT}, the approach is given the name \emph{\acf{MoDT}}. Our method builds on top of the one by Vasic et al. \cite{Vasic2019} who use a similar architecture, however, do not utilize its potential for interpretability. The main contributions of \ac{MoDT} are a novel interpretable gating function that can reduce a multi-dimensional gating problem to a two-dimensional one, the usage of a fast linear regression for finding promising model parameters, and a visualization method that matches DTs and gating function. Thereby, \ac{MoDT} is the first modification of the \ac{MoE} architecture using \acp{DT} that is fully optimized towards interpretability.

In the following, fundamental techniques and a problem setting for MoDT are introduced. The developed method is described in detail in Sec.~\ref{sec:modt}. Next, \ac{MoDT} is put onto the test bench with an experimental study that investigates its general performance boundaries. In Sec.~\ref{sec:discussion}, the implications of this experiment and the usage of \ac{MoDT} as a method of \ac{IML} are discussed. Finally, the paper is briefly summarized.


\section{Problem Formulation}
\label{sec:problem}


In the following, it is assumed that a training dataset $\SD$ is given comprising pairs $(x_i, y_i)$, with $i=1\ldots n$, where $x_i$ is the feature vector (extended by an element with value one) of dimension $p + 1$ 
and $y_i$ is the output (class label). All feature vectors are stored in a matrix $\X = [x_1\ x_2 \ldots x_n]\T$ and $y = [y_1\ y_2\ldots y_n]\T$ is the target vector comprising all outputs. Then, a supervised classification model tries to learn the relationship between $x$ and $y$ such that it is able to predict $y'$ for vector $x'$ of a potentially different dataset~$\SD'$.

A \ac{DT} is a supervised \ac{ML} method that decomposes a problem into a series of decisions. This decomposition resembles the human decision process. Therefore, \acp{DT} are considered to be more accessible to humans than other techniques of ML and often given as a prime example of an \ac{IML} method \cite{Bishop2006, Molnar2019}.
Yet, while small trees are considered to be very interpretable, \acp{DT} can quickly become hard to understand if too many nodes (decision splits) are included. 
Therefore, the complexity of a \ac{DT} is typically limited when interpretability is desired. This can be achieved by limiting the maximum tree depth $d$.

\begin{figure}[t]
    \includegraphics[width=0.49\textwidth]{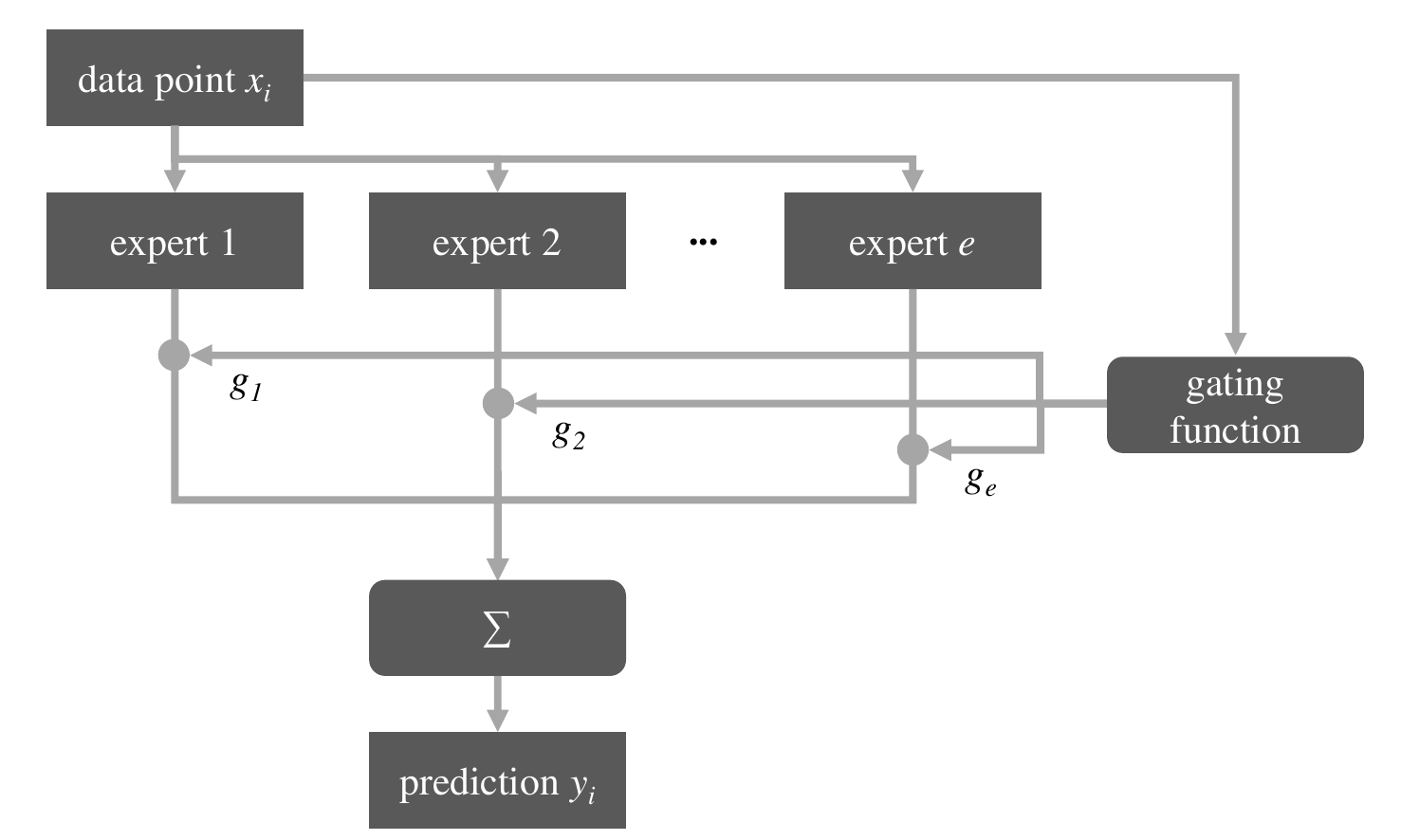}
    \caption{Architecture of an \ac{MoE} prediction model for a single data point $x_i$. Depending on that point $x_i$, the gating function outputs weights which are used to aggregate the expert's individual predictions.}
    \label{fig:moe}
\end{figure}

In 1991, the original \ac{MoE} model was introduced by Jacobs et al. \cite{Jacobs1991}. The general idea behind MoE is that multiple smaller problems are easier to solve than a single larger one. \ac{MoE} can be seen as an ensemble method that uses multiple submodels (called \emph{experts}) to solve a supervised \ac{ML} problem. The number of experts $e$ needs to be set in advance. Then, each expert is trained independently on a subset of the input space. This is beneficial if the subsets incorporate different input-output relationships such that the experts can learn these distinct patterns. In \ac{MoE}, the assignment of the subsets to the experts is conducted by a so-called \emph{gating function}, which itself is subject to optimization. The gating function can be any function that maps an input data point to a subset of the input space. It is optimized alternately with the experts in an iterative fashion using the \ac{EM} algorithm \cite{Bishop2006}. Thereby, a ``chicken and egg'' dilemma can be avoided as the optimization of the gates depends on the optimization of the experts and vice versa.
After training, when an \ac{MoE} model is used for prediction, the gating function decides which expert is used for the prediction of a data point. Here, it is possible to only use the expert with the highest gating value or, alternatively, multiple experts whose outputs are combined using weights provided by the gating function. 
Figure \ref{fig:moe} depicts the architecture of a trained MoE model.

For an input data point $x_i$,
the corresponding prediction $y_i$
of the MoE model is given by
$y_i = \sum_{j=1}^{e}g_{j} \cdot z_{j}$  
where $e$ denotes the number of experts, $g_j$ the gating weight for expert $j$, and $z_j$ the prediction of expert $j$ \cite{Bishop2006}. It holds that $g_j \in [0,1]$ and $\sum_{j=1}^{e}g_{j} = 1$ \cite{Bishop2006}. This is often achieved through the usage of the softmax function. In the case where only one expert is used for the overall prediction, one weight is set to one while the others are set to zero. This is mostly a necessity for interpretability, as a combination of multiple experts can be difficult to  understand for humans. Another necessity is that the gating function must be interpretable. Common MoE gating function choices like artificial neural networks are thus not appropriate for \ac{IML}.

\ac{MoDT} is an instance of \ac{MoE} where a linear model is used as gating function and \acp{DT} are used as experts.
A known problem of \acp{DT} is their inability to efficiently capture linear relationships due to the decision splits being always aligned with the dimensions of the input space. If two classes are optimally separated by a 45$^\circ$ boundary with respect to two dimensions, a \ac{DT} can only approximate this with a large number of splits \cite{Bishop2006}, which is not suitable in the context of interpretability. This problem can be addressed by MoDT by employing a linear gating function in conjunction with DTs. 

\section{Mixture of Decision Trees}
\label{sec:modt}
The key components of \ac{MoDT} are introduced in the following. In Sec.~\ref{sec:modt_overview} an overview of the entire \ac{MoDT} architecture is given. Sec.~\ref{sec:modt_training} describes the training based on the \ac{EM} algorithm, while Sec.~\ref{sec:modt_prediction} briefly addresses the prediction of the output $y$ given a trained \ac{MoDT} and an input vector $x$. The visualization capabilities of \ac{MoDT} to support interpretability are demonstrated in Sec.~\ref{sec:modt_visualization} by means of a toy dataset.

\subsection{Overview of the MoDT Architecture}
\label{sec:modt_overview}
The proposed method \ac{MoDT} 
can be categorized as an ensemble method that uses multiple \acp{DT} as predictors. To facilitate interpretability, only a small single-digit number of \acp{DT} should be used. Unlike many other ensemble methods like \acp{RF} that combine the predictions of multiple DTs, MoDT only uses a \emph{single} \ac{DT} for a \emph{single} prediction. Again, this is motivated by the requirement of interpretability: A single DT is usually considered to be interpretable while an aggregation of multiple \acp{DT} is not. 


The number of regions and thus \acp{DT} $e$ has to be specified by the user. Alternatively, our implementation also includes methods to estimate a suitable number, e.g., based on Gaussian mixture models \cite{Bishop2006}. Each region is associated with a distinct \ac{DT} that is trained on the subset of $\X$ that falls into the region. The core idea is that the regions contain distinct patterns that can be represented best by distinct \acp{DT}. To allow interpretability, the complexity of the \acp{DT} needs to be limited, which is achieved by setting a maximum \ac{DT} depth $d$.

The gating parameter matrix $\mat \theta \in \NewR^{(p+1) \times e}$ defines how the gating function splits the input space into regions and is initialized randomly. In \ac{MoDT}, the gating function is a linear model including a bias term (intercept), which explains the additional dimension in $x_i$. 
The $j$-th column of $\theta$ comprises the parameters of the linear gating function corresponding to the $j$-th DT, with $j = 1,\ldots, e$. 
The outputs of the gating function, called \emph{gating values} in the following, determine which parts of the input data are used for the training of each \ac{DT}. 
The matrix of gating values $\G\in \NewR^{n\times e}$ is the result of a softmax on a matrix multiplication using the gating parameters $\theta$ and the input data $\X$ according to 
\vspace{-.5mm}
\begin{align}
\label{eq:gating}
\G = g(\X, \theta) 
&= \begin{bmatrix}
\nicefrac{1}{\lvert a_1 \lvert} & \nicefrac{1}{\lvert a_1 \lvert} & \cdots & \nicefrac{1}{\lvert a_1 \lvert} \\
\vdots &\vdots & &\vdots\\
\nicefrac{1}{\lvert a_n \lvert} & \nicefrac{1}{\lvert a_n \lvert} & \cdots & \nicefrac{1}{\lvert a_n \lvert}
\end{bmatrix}
\circ \A~,
\end{align}
where $\A = [a_1\ a_2\ \ldots\ a_n]\T = \exp(\X \cdot \theta ~- ~\max( \X \cdot \theta))$, $a_i$ is the $i$-th row of $\A$, $\circ$ is the Hadamard (element-wise) product, and 
$g(.)$ is the \emph{gating function}. Hence, each row of $\G$ sums to one. 
The calculation on the right-hand side of Eq.~\eqref{eq:gating} corresponds to a row-wise application of a variation of the softmax function, where in each row of $\X \cdot \theta$ the row-wise maximum of $\X\cdot \theta$ is substracted inside the exponential function. 
The resulting gating values $\G$ determine the allocation of each data point to the regions. One element $g_{ij}$ of $\G$ represents the probability of choosing expert $j$ for data point $i$.

For MoDT, it is possible to use a gating function that either considers all $p$ features of $\X$, or alternatively, just two features.
Due to their linear nature, both variants behave relatively predictable. Yet, the \ac{2D} gating function is beneficial for interpretability as it has the great advantage of being comprehensibly visualizable using a \ac{2D} plot. For the \ac{2D} gating function, the question arises, which two out of all $p$ features should be selected for the gating function. In the implementation of \ac{MoDT}, multiple feature importance methods (for example based on \acp{DT}, linear models, or principal component analysis) are included for selecting promising candidates. From an interpretability perspective, users can also manually select two features that are well known to them. 
The reduced dimension in case of a \ac{2D} gating function affects the size of $\theta$ and only the two selected feature columns of $\X$ are considerede for calculating $\A$ in \eqref{eq:gating}.
In our experiments (see Section~\ref{sec:experiments}), it will be tested how the limitation to two features impacts performance in comparison to the ``full''
gating function using all features.

The ``goodness'' of a region depends on the ``goodness'' of the associated \ac{DT} and vice versa. To avoid a ``chicken and egg'' dilemma, the training algorithm thus uses the two-step EM algorithm, which alternately updates the gating function and the \acp{DT}. More precisely, during the E-step of the EM algorithm, a \ac{DT} is trained for each region. 
Technically, this is achieved by using the output vector of the gating function for each data point $x$ as weight vector during the training of the \acp{DT}. 
Then, the expected ``goodness'' of the current combination of \ac{DT} and the gating parameters $\theta$ with respect to the true prediction targets $y$ is calculated. This expectation is used in the M-step to optimize $\theta$ and the gating function, respectively. The E-step and M-step are repeated until a stopping criterion or a fixed number of iterations is reached. 

\subsection{Training}

In contrast to other \ac{MoE} architectures, where the training of the experts is conducted in the M-step \cite{Yang2009, Bishop2006}, we prefer training the experts during the E-step.
Further, regardless of the choice of the 2D or the full gating function, the following training procedure is the same.

During the training, the complexity of the \acp{DT} is controlled by a hyperparameter $d$ that limits the maximum depth of the \acp{DT}. There are also other possibilities to limit the complexity of a \ac{DT} (like the number of leaf nodes), however here, the maximum depth is preferred as it potentially reduces the variance of the \acp{DT} between iterations, resulting in a more stable training process. 

Once the \acp{DT} are trained, the expectation $E$, which gives the E-step its name, can be calculated. For this purpose, the gating function and a so-called \emph{confidence function}, denoted as $c(\ST, \X, y)$, are necessary, where $\ST$ is the set of all \acp{DT}. At first, the gating values $\G$ are calculated by means of Eq.~\eqref{eq:gating}. 
Secondly, the confidence function $c(\ST, \X, y)$ outputs the \acp{DT}' confidences for the correct predictions. For this purpose, as a first step, all trained \acp{DT} $z_j\in \ST$ perform predictions for the input $\X$. Instead of assigning a particular class to a data point $x_i$, each DT $z_j$ calculates a vector comprising the prediction probabilities for all possible classes of $x_i$. 
This vector and the true classes $y$ are then used to extract the probability for a correct prediction of each \ac{DT} for each data point of $\X$. Thus, the output of $c(.)$ corresponding to one data point $x_i$ is the vector $c_i$, where the $j$-th element of $c_i$ corresponds to the probability of DT $z_j$ outputting the true class $y_i$ of input vector $x_i$. Finally, the expectation $\mE$ can be calculated according to
\vspace{-.5mm}
\begin{equation} 
\label{eq:E}
\mE = \begin{bmatrix}
\nicefrac{1}{\lvert b_1 \lvert} & \nicefrac{1}{\lvert b_1 \lvert} & \cdots & \nicefrac{1}{\lvert b_1 \lvert} \\
\vdots &\vdots & &\vdots\\
\nicefrac{1}{\lvert b_n \lvert} & \nicefrac{1}{\lvert b_n \lvert} & \cdots & \nicefrac{1}{\lvert b_n \lvert}
\end{bmatrix}
\circ \B~,
\end{equation}
where $\B = [b_1\ b_2\ \ldots \ b_n]\T = \G\circ\C$ with $\C = [c_0\ c_1\ \ldots\ c_e] \in \NewR^{n\times e}$ being the output matrix of function $c(\ST, \X, y)$. Each row of $\mE$ sums to one. 

The calculation of one row of $\mE$ in Eq.~\eqref{eq:E} is illustrated in a numeric example: Given an \ac{MoDT} model with three experts, assuming the gating values for a data point being $[0.7\ 0.2\ 0.1]$, and the experts' probabilities for the correct prediction are given by $[1.0\ 0.9\ 0.5]$, the expectation for this data point is $[0.70\ 0.18\ 0.05] ~ / ~ 0.93 = [0.753\ 0.194\ 0.054]$.
If these values were directly interpreted as new gating values, it can be noticed that the probability for the best-performing expert increases, while the opposite is true for the weaker experts. This observation is crucial for the training algorithm of \ac{MoDT}.

\label{sec:modt_training}
\begin{figure}[!t]
    \centering
    \includegraphics[width=0.925\columnwidth]{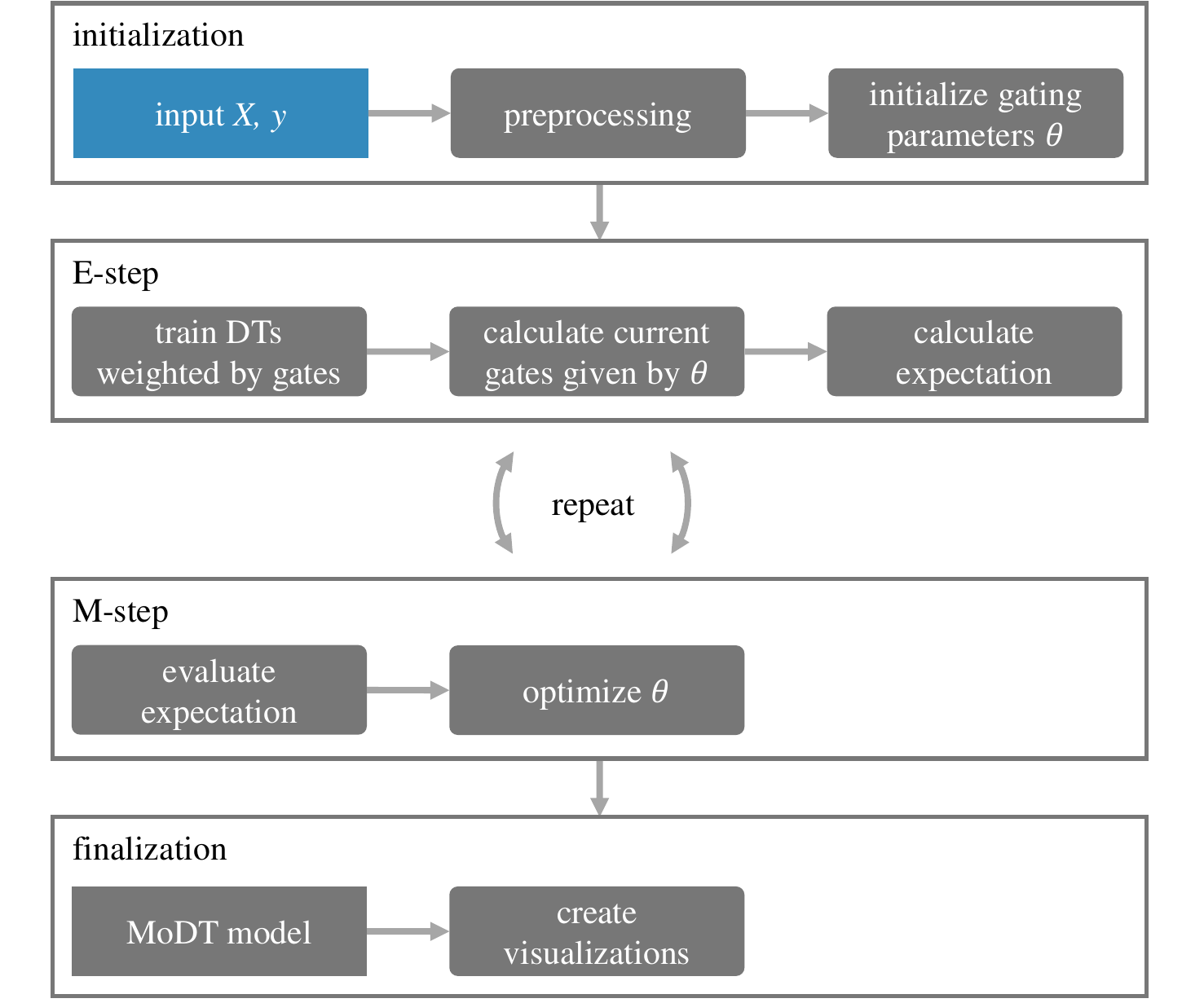}
    \caption{Overview of the training process of MoDT.}
    \label{fig:training}
\end{figure}

The newly calculated expectation does not replace the gating values but is used as a basis for a controllable optimization during the M-step of the \ac{EM} algorithm. The original gating values and the expectation are used to calculate an optimization direction similar to a gradient that is applied to the gating parameters $\theta$, governed by an adjustable learning rate parameter. In classic \ac{MoE} architectures, the expectation is used within a log-likelihood function that then serves as the loss for iterative and possibly computationally expensive optimization algorithms \cite{Bishop2006, Yang2009}. However, an experiment by Yang and Ma \cite{Yang2009} compares multiple optimization algorithms and concludes that a rather simple least squares linear regression is favorable for an \ac{MoE} architecture with a linear gating function. 
Therefore, \ac{MoDT} primarily uses a linear regression for the update of the gating parameters $\theta$ which does not require a gradient descent. The linear regression creates a linear model with coefficients $\beta$ that fits $\X$ to a target. For this purpose, the \ac{RSS}, i.e., the sum of squared deviations of the predicted targets to the correct targets is minimized. The target of the regression is not the earlier calculated expectation $\mE$, but the difference of $\mE$ and the gating values $\G$. This target $\mE-\G \in \NewR^{n \times e}$ can be interpreted as the direction towards the desired new gating values. 
Formally, the coefficients $\beta$ are chosen to minimize
\begin{equation} 
\text{RSS}(\beta) = \sum_{i=1}^{n}\big((e_i-g_i)-x_i\T\beta\big)\cdot \big((e_i-g_i)-x_i\T\beta\big)\T~,
\end{equation}
with $e_i$ and $g_i$ being the $i$-th row of $\mE$ and $\G$
, respectively~\cite{Hastie2009}. 
The resulting coefficients $\beta \in \NewR^{(p+1) \times e}$ 
can be seen as the internals of a linear model that matches $\X$ to $\mE-\G$, similar to the linear gating function of MoDT which matches $\X$ to~$\G$.

To calculate the new gating parameters $\theta_\mathrm{new}$, the update equation
$\theta_\mathrm{new} = \theta_\mathrm{old} + \gamma\cdot \beta$
%
is employed, with $\theta_\mathrm{old}$ being the current gating parameters and $\gamma > 0$ being the learning rate. 
The update of $\theta$ concludes the M-step. 

Once the EM algorithm has terminated, the trained MoDT model can be used to create visualizations for both the gating function (if the 2D variant is selected) and the DTs.
An overview of the training process is given in Figure \ref{fig:training}.


\subsection{Prediction}
\label{sec:modt_prediction}
The prediction process takes unseen data $\X'$ as input of the gating function which outputs the gating values $\G$. It is important to note that the earlier outlined calculation of $\G$\ (see Eq.~\eqref{eq:gating}) is only used for the training and the gating function for the prediction is slightly different. In particular, the gating function for the prediction will only output the expert (\ac{DT}) with the highest probability. 
The final prediction $y_i$ for a data point $x_i$ is thus given by
\vspace{-.5mm}
\begin{equation} 
\begin{split}
         y_i & = \sum_{j=1}^{e} z_j(x_{i}) \cdot \mathds{1}(j = \argmax{k}(g_{ik}))~,\\
\end{split}
\end{equation}
where $z_j$ is the prediction of the $j$-th DT and $\mathds{1}$ is the indicator function. 
The indicator function returns one if a proposition is true, i.e., if data point $x_i$ belongs to expert $e_j$. Otherwise, zero is returned.

In contrast, it would also be possible to use the probabilities of the experts in place of the argmax function. In this case, the final prediction would be the result of a weighted sum of the predictions of all \acp{DT}. This architecture would resemble the common structure of \ac{MoE} as seen earlier in Figure~\ref{fig:moe}. This setup is also used in the paper of Vasic et al. \cite{Vasic2019} where it is called soft-gate prediction. However, for \ac{MoDT}, such a combination of \acp{DT} is avoided as the increased complexity is unsuited for interpretability. 


\subsection{Visualization and Application Example}
\label{sec:modt_visualization}

\begin{figure}[!t]
    \centering
    \includegraphics[width=0.8\columnwidth]{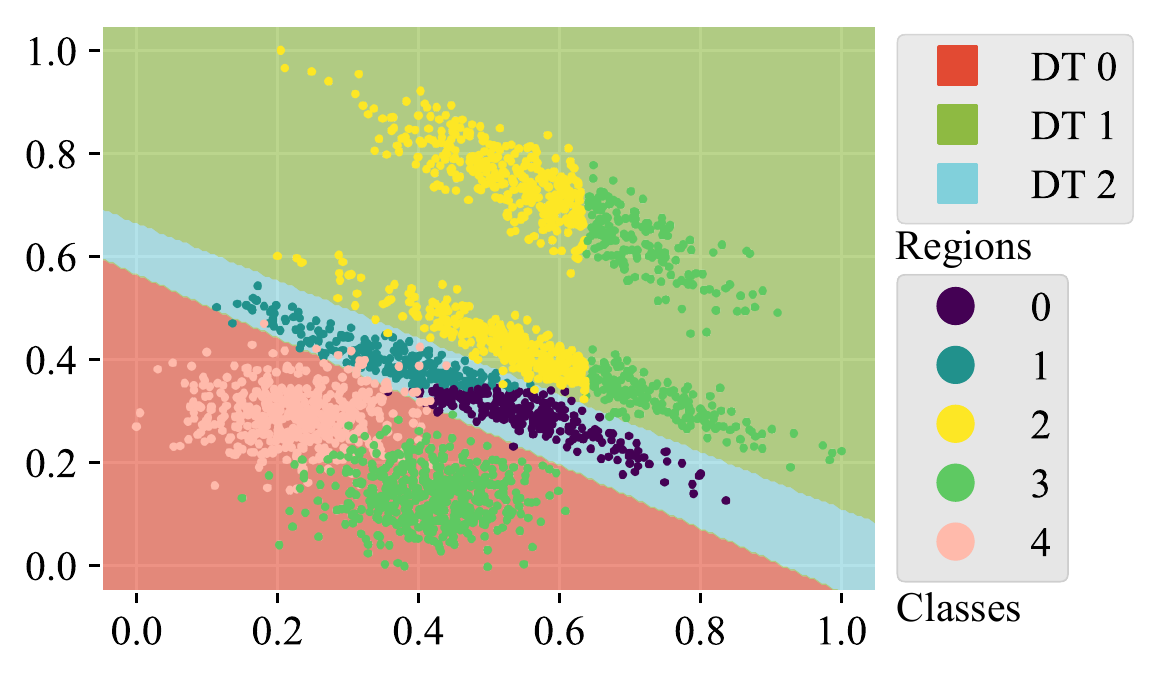}
    \caption{Exemplary decision area of the 2D gating function.}
    \label{fig:gate_example0}
		\vspace{-2mm}
\end{figure}

In order to allow interpretability, a reasonable and comprehensive visualization is indispensable.
If the feature space is \acl{2D} or the \ac{2D} gating function is used, the gating function and the resulting regions can be plotted as depicted in Figure~\ref{fig:gate_example0}.
Given the plot in Figure \ref{fig:gate_example0}, a user can understand that \ac{MoDT} solves the classification problem by dividing the dataset into three regions that incorporate three distinct patterns. A legend that enumerates the regions/\acp{DT} (Figure \ref{fig:gate_example0}, top right) and a legend for the classes of the data points (Figure \ref{fig:gate_example0}, bottom right) is added to the plot. The colors are chosen such that the data points remain visible on top of the regions. 
\begin{figure}[!t]
        \centering
				\includegraphics[width=135pt]{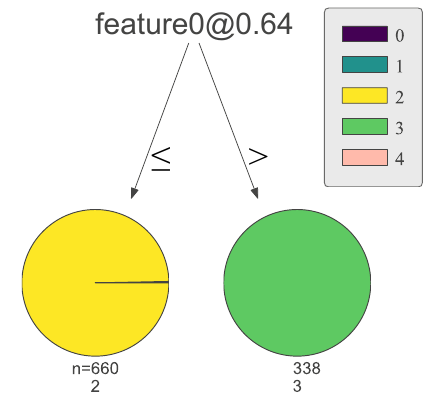} 
				\vspace{-1mm}
        \caption{DT corresponding to the green region in Figure \ref{fig:gate_example0}.}
        \label{fig:dt_example0}
\end{figure}
One of the three \acp{DT}, i.e., the \ac{DT} for the top-most green region can be seen in Figure \ref{fig:dt_example0}. Here, the region is expressed by a simplistic \ac{DT} with a single decision criterion, which often is called a decision stump. The top-right legend corresponds to the classes of the classification problem. The \acp{DT} for the blue and red regions are not depicted.

\begin{figure}[!t]
    \centering
    \includegraphics[width=\columnwidth]{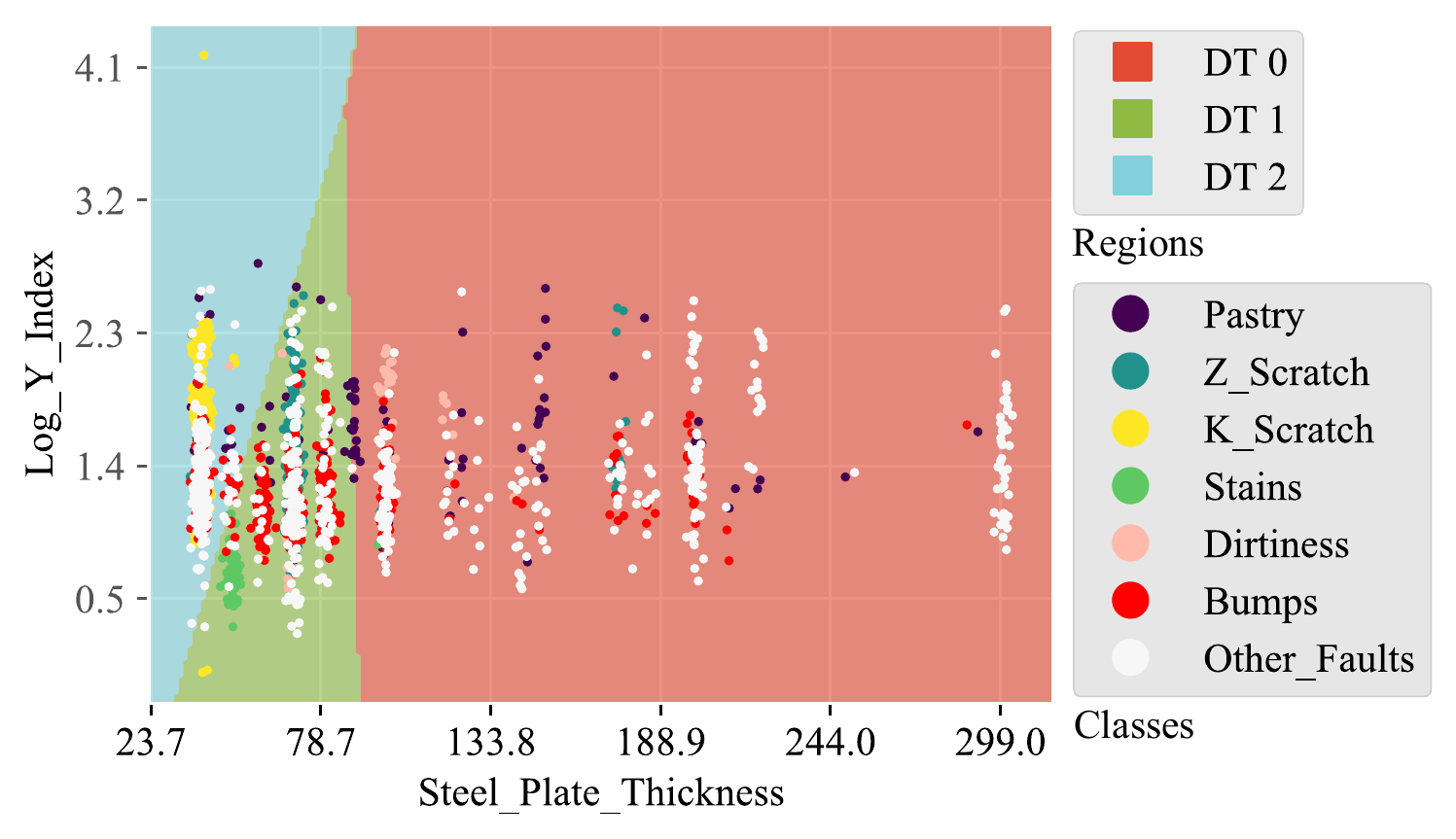}
    \caption{Decision area of the 2D gating function on the steel dataset.}
    \label{fig:steel_gate_d2}
    \vspace{-2mm}
\end{figure}

\begin{figure}[!t]
    \centering
    \includegraphics[width=\columnwidth]{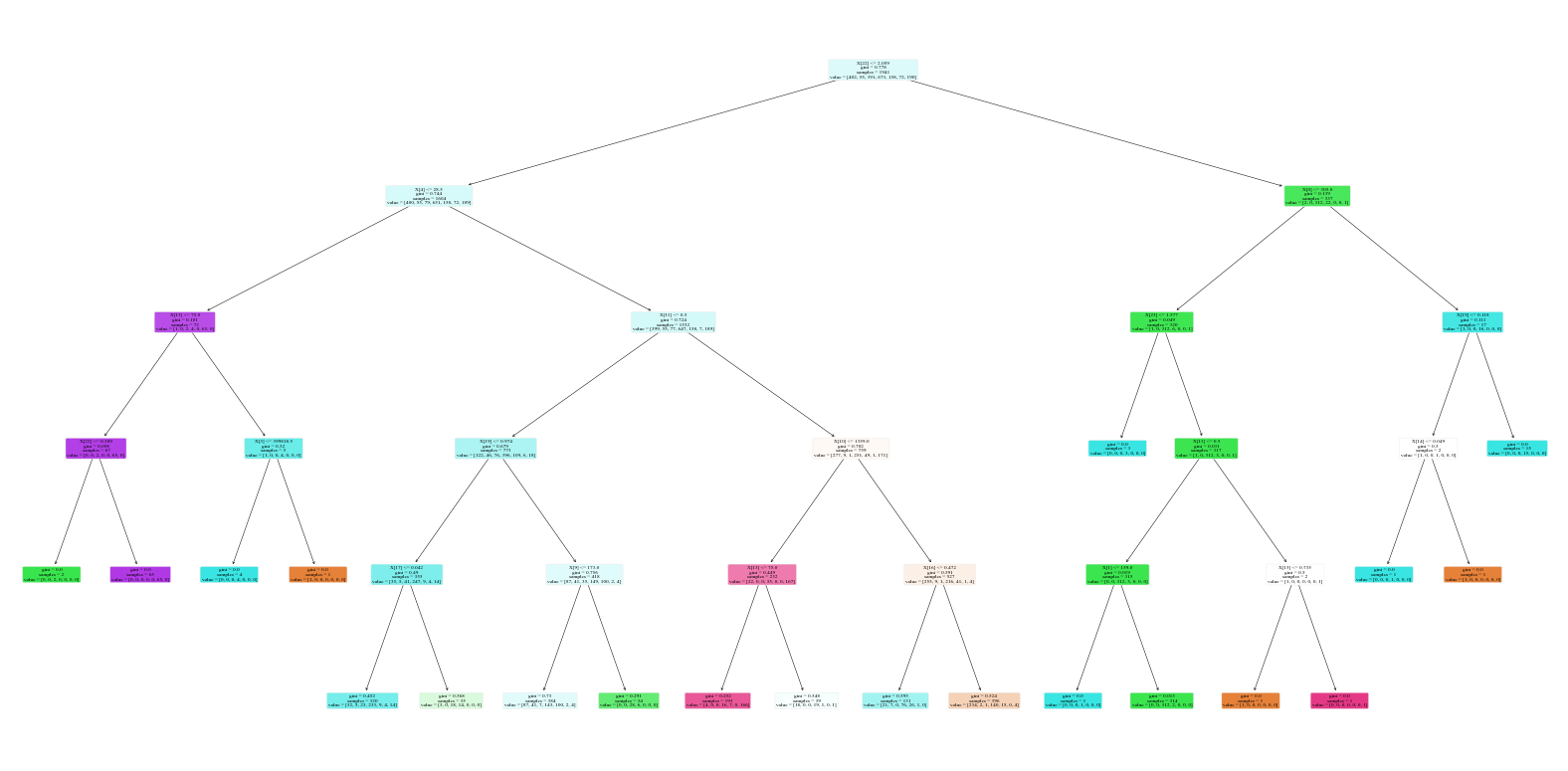}
    \caption{An example of a classic DT that is not very interpretable.}
    \label{fig:bad_example}
\end{figure}

An example of the visualization of the 2D gating function on a more complex problem, namely the steel dataset\cite{Dua2019}, is depicted in Figure \ref{fig:steel_gate_d2}. The two features \texttt{Log\_Y\_Index} and \texttt{Steel\_Plate\_Thickness} have been selected using feature importance. 
Again, three experts are used. Although the corresponding \acp{DT} are not depicted in this paper, it can be seen how the regions incorporate distinct behavior. On this dataset, a MoDT model can reach a training accuracy of around $68\%$ when three DTs with a maximum depth of two are used (see Chapter \ref{sec:experiments_results}). Here, MoDT uses at most $3 \cdot 7 = 21$ nodes (a DT of depth two has a maximum of seven nodes). For comparison, as a negative example, Figure \ref{fig:bad_example} depicts a single DT that reaches the same accuracy. It uses almost twice the number of nodes (39) and---in contrast to MoDT---is too large to be conveniently plottable and interpretable. This dataset is revisited also in the experiments in Sec.~\ref{sec:experiments}.

For visualizing \acp{DT}, many software packages are freely available. Although the Python library dtreeviz (\url{https://github.com/parrt/dtreeviz}) does not check all interpretability boxes, it is a good match for \ac{MoDT}. A great advantage of the library is that the colors of the classes can be adjusted with ease. This way, the colors of the \ac{DT} correspond to the colors of the classes in the gating plot, thereby leveraging the joint comprehensibility of both graphics. 


\section{Experiments}
\label{sec:experiments}

\begin{table*}[t!]
\caption{Datasets used in the experiment.}
\vspace{-1mm}
\label{tab:datasets}
\begin{center}
\begin{tabular}{ |l|r|r|r|r|r|l| } 
\hline
\multicolumn{1}{|l|}{dataset} & \multicolumn{1}{l|}{data points} & \multicolumn{1}{l|}{classes} & \multicolumn{3}{c|}{features (num./cat./enc.)} & \multicolumn{1}{l|}{description} \\
\hline\hline
abalone & 4,177 & 3 & 7 & 1 & 10 & Predict age group of abalones (sea animals) based on physical measurements. \\
\hline
banknote & 1,372 & 2 & 4 & 0 & 4 &  Predict authenticity of banknotes.\\
\hline
breast & 569 & 2 & 10 & 0 & 10 & Predict breast cancer based on properties extracted from images. \\
\hline
cars & 1,728 & 4 & 0 & 6 & 21 & Predict quality of cars based on  categorical features only. \\
\hline
contraceptive & 1,473 & 3 & 2 & 7 & 24 & Predict contraceptive method. \\
\hline
iris & 150 & 3 & 4 & 0 & 4 & Predict species of a flower. Popular dataset. \\
\hline
steel & 1,941 & 7 & 27 & 0 & 27 & Predict defects of steel plates.\\
\hline
students & 666 & 4 & 0 & 11 & 49 & Predict performances of students on an entrance examination. \\
\hline
\end{tabular}
\end{center}
\end{table*}

In the following, the general performance of \ac{MoDT} in terms of classification accuracy is tested and compared against itself and other classic tree-based \ac{ML} models with varying complexity. MoDT is also compared to non-interpretable ML methods which generally should not be used for typical \ac{IML} scenarios in which safety and ethical standards must be guaranteed \cite{Rudin}. However, as these methods are typically way more powerful regarding pure predictive performance, it can already be seen as a success if an interpretable method comes close to the performance of a non-interpretable method.


\subsection{Experimental Setup}
\label{sec:experiments_setup}

To estimate an upper bound of \ac{MoDT}'s performance, a hyperparameter optimization algorithm is used that aims at finding a well-performing combination of hyperparameters. 
For this, the Python library Optuna \cite{optuna2019} is utilized. 
The experiment is conducted with eight publicly available datasets 
\cite{Dua2019}, summarized in Table \ref{tab:datasets}. 
As some of the datasets contain categorical features that are not supported by the scikit-learn \ac{DT} implementation, MoDT uses a one-hot-encoding for categorical features. 

Each dataset is shuffled into training data (75\%) and test data (25\%). Then, the experiment is conducted independently for each dataset. Each training dataset is firstly used for the hyperparameter search which is repeated 500 times.
Instead of choosing the single best hyperparameter combination, the ten best combinations are identified. These are then used to train ten distinct \ac{MoDT} models on the training data. Lastly, the final performance per combination is measured with 100 repetitions on the held-out 25\% test data. In addition, the standard deviation and the training accuracy are denoted. The idea is that, by using the ten best combinations, the results are not overly optimistic but represent a reasonable performance that could be reproduced with an average hyperparameter search. While it is theoretically possible that the found hyperparameters are biased towards the 75\% training data, the evaluation on the independent test data should not yield over-optimistic results.

MoDT is used with three experts ($e=3$) and a DT depth of two ($d=2$). Derived from earlier experiments, this configuration seems to be an appropriate compromise between expressiveness and simplicity.
Since \acp{DT} are well-known for their interpretability, and \ac{MoDT} builds upon them, \ac{MoDT} is compared to \acp{DT} with a maximum depth $d$ between two and four. Since $d=4$ can already be problematic for interpretability (see Section \ref{sec:discussion}), a greater depth is not tested.
As the Python library scikit-learn \cite{scikit-learn} is used for the DTs within the implementation of \ac{MoDT}, the \acp{DT} for the comparison are also created with scikit-learn. 
Furthermore, \ac{MoDT} is compared to \acp{RF}, which are valued for their good performance \cite{Molnar2019} and are also based on \acp{DT}. Although \acp{RF} are not interpretable, the comparison allows assessing \ac{MoDT} from a pure performance perspective. Firstly, \acp{RF} are used with a configuration of $d=2$ and $e=3$ that is identical to the here used configuration of \ac{MoDT}. Secondly, the default configuration of scikit-learn is used, which employs 100 \acp{DT} and imposes no restriction on the depth. The accuracy is evaluated using a 75/25\% training/test split and 100~repetitions. 

To allow a fair comparison, the same preprocessing steps that are included in \ac{MoDT} are also applied to the training data of the alternatives. However, no extensive hyperparameter searches are conducted for the \acp{DT} and random forests. Yet, it is assumed that the default scikit-learn hyperparameters already yield reasonable results, especially in the deployed low complexity configurations. In addition, it would be problematic to compare the default \acp{DT} within the MoDT ensemble with optimized individual \acp{DT}. 
The complete above process is firstly performed for an \ac{MoDT} configuration using the assumingly more interpretable but less powerful \ac{2D} gating function, and secondly for the \ac{FG} function.

\subsection{Results}
\label{sec:experiments_results}

\begin{table*}[tbh]
\caption{Performance of \ac{MoDT} with \ac{FG} and \ac{2D} gate, \acp{DT} and RFs. Best interpretable method: grey background, best overall: bold.} 
\begin{center}
\begin{tabular}{ |r|r|r|r|r|r|r|r|r|r| } 
\cline{2-10}
\multicolumn{1}{c|}{} & \multicolumn{2}{c|}{\ac{MoDT} 2D d$=$2 e$=$3} & \multicolumn{2}{c|}{\ac{MoDT} FG d$=$2 e$=$3} & \multicolumn{3}{c|}{\ac{DT}} & \multicolumn{2}{c|}{\ac{RF}} \\
\hline
dataset & training & test & training & test & d$=$2 &$d=3$ & $d=4$ & d$=$2 e$=$3 & d$=$*$~$e$=$100  \\
\hline
abalone & .74$\pm$.00 & .71$\pm$.01 & .75$\pm$.01 & \cellcolor{custom-grey}\textbf{.73$\pm$.01} & .67$\pm$.01 & .69$\pm$.01 & .70$\pm$.01 & .67$\pm$.01 & \textbf{.73$\pm$.01} \\
banknote & 1.00$\pm$.00 & .99$\pm$.00 & 1.00$\pm$.00 & \cellcolor{custom-grey}\textbf{1.00$\pm$.00} &  .91$\pm$.02 & .93$\pm$.01 & .95$\pm$.02 & .89$\pm$.04 & .99$\pm$.00 \\
breast & .96$\pm$.00 & .94$\pm$.01 & .97$\pm$.01 & \cellcolor{custom-grey}\textbf{.95$\pm$.02} &  .91$\pm$.02 & .92$\pm$.02 & .92$\pm$.02 & .91$\pm$.02 & .94$\pm$.02 \\
cars & .82$\pm$.01 & .78$\pm$.01 & .92$\pm$.01 & \cellcolor{custom-grey}.88$\pm$.01 &  .78$\pm$.02 & .79$\pm$.01 & .81$\pm$.02 & .71$\pm$.02 &\textbf{ .96$\pm$.01} \\
contraceptive & .59$\pm$.01 & \cellcolor{custom-grey}\textbf{.58$\pm$.01} & .57$\pm$.01 & .53$\pm$.02 &  .48$\pm$.02 & .52$\pm$.03 & .55$\pm$.02 & .46$\pm$.04 & .52$\pm$.02 \\
iris & .99$\pm$.01 & .95$\pm$.02 & .99$\pm$.01 & \cellcolor{custom-grey}\textbf{.96$\pm$.02} &  .94$\pm$.03 & .94$\pm$.04 & .94$\pm$.03 & .92$\pm$.06 & .95$\pm$.03 \\
steel & .68$\pm$.01 & \cellcolor{custom-grey}.68$\pm$.01 & .70$\pm$.02 & .67$\pm$.01 &  .53$\pm$.02 & .53$\pm$.02 & .61$\pm$.02 & .53$\pm$.02 & \textbf{.78$\pm$.02} \\
students & .58$\pm$.02 & .45$\pm$.01 & .53$\pm$.01 & .41$\pm$.03 &  .48$\pm$.04 &\cellcolor{custom-grey} \textbf{.51$\pm$.03} & .50$\pm$.04 & .44$\pm$.06 & .49$\pm$.03 \\
\hline
\end{tabular}
\label{tab:performance_all}
\end{center}
\end{table*}

The average accuracies and the standard deviations using the \ac{MoDT} classifier model with the \ac{2D} and \ac{FG} function are erported in Table \ref{tab:performance_all}. The best-performing interpretable methods on test accuracy (without consideration of the standard deviation) are highlighted with a grey background. \acp{RF} (as they are not interpretable) and the training accuracy of \ac{MoDT} are not considered for this highlighting. The best overall methods on test accuracy are highlighted in bold. 

In comparison to \acp{DT} with $d$ up to four, \ac{MoDT} using a 2D gating function achieves a better test accuracy on 6 out of 8 datasets, hereby 
neglecting the standard deviations. In all cases, the test accuracy is close to the training accuracy (within 1--4 percentage points) except for the students dataset, where the test accuracy is significantly worse than the training accuracy. On this dataset, the \acp{DT} collectively outperform \ac{MoDT}. 
In comparison with \ac{MoDT} using the 2D gating function, the \acp{RF}' configuration with the same complexity as \ac{MoDT} achieves lower scores on all datasets. Some of the differences are relatively large, e.g., \ac{MoDT} performs 15 percentage points better on the steel dataset. \ac{MoDT} achieves equal or better performance than the significantly more complex \ac{RF} configuration with 100 \acp{DT} in 4 of the 8 datasets.

The full gating function exploits all features of $\X$ instead of just two. Is therefore can be more expressive but it is less interpretable. Therefore, it should be investigated, how this variant compares to the 2D gating variant.
Using the full gating function, \ac{MoDT} achieves a higher performance than the DTs in 6 out of 8 datasets. In contrast to the \ac{2D} variant, \ac{MoDT} using a full gate now outperforms the \acp{DT} on the cars dataset. The opposite is observable for the contraceptive dataset, here, a \ac{DT} with $d=4$ now beats the performance of \ac{MoDT} with a full gate. The full gate variant of \ac{MoDT} achieves equal or better performance than the random forests configuration with 100 \acp{DT} in 5 of the 8 datasets. 

Although the full gating function is in theory more powerful than its \ac{2D} counterpart, a direct comparison suggests that the \acl{FG} variant is not universally better. In 5 of the 8 datasets, both variants achieve roughly similar performance ($\pm2$ percentage points). The \acl{FG} variant achieves a 10 percentage points better performance on the cars dataset. Conversely, the contraceptive and students datasets benefit from the \ac{2D} gate variant. Here, the test results are 5 and 4, respectively, percentage points higher.
The training time of both variants is almost identical and ranges between $0.1\,$s for the iris dataset and $1.4\,$s for the abalone dataset.

\section{Discussion}
\label{sec:discussion}

The comparison with interpretable \acp{DT} and \acp{RF} shows that \ac{MoDT} is working as intended and can provide a benefit over individual \acp{DT}. On average, \ac{MoDT} clearly outperforms single \acp{DT} with $d=2$.  \ac{MoDT} continues to outperform \acp{DT} with $d=4$, which are arguably already rather complex and can be hard to interpret. A DT of $d=4$ can use more nodes (31) than \ac{MoDT} with $d=2$ and $e=3$ (21). Therefore, it can be assumed that the gating function works as intended and cannot be replaced by a small number of additional \ac{DT} splits.

On average, \ac{MoDT} is clearly superior to a \ac{RF} with the same complexity.  
As expected, the \ac{RF} variant with a vastly more complex structure mostly outperforms \ac{MoDT}. However, the resulting \ac{RF}s are also not interpretable.
Therefore, considering the way simpler structure of \ac{MoDT}, it can be seen as a success that an interpretable method can achieve similar and sometimes even higher scores.


The comparison of the 2D gate and the full gate reveals that in all but one cases, namely the cars dataset,
a 2D gating function is sufficient. This is an important result as the 2D gate is highly beneficial for interpretability. The variant can achieve similar and sometimes even better performance than the full gate. Possibly, the reduced complexity can prevent overfitting. However, the fixed number of iterations used in the experiment might be disadvantageous for the \acl{FG} variant. Since the variant is more complex, there are possibly more directions for optimization, and the training algorithm might need more iterations to explore them.

The most important hyperparameters of \ac{MoDT} are the maximum depth $d$ of the \acp{DT} and the number of experts $e$. While it might seem difficult to set these values, a conceptional analysis of interpretability suggests a rather small range of suitable values. In one of the most cited papers in the field of psychology \cite{Miller1956}, the hypothesis known as \emph{Miller's law} is introduced, which says that the average human can only keep approximately seven different chunks in their working memory. According to this, a binary \ac{DT} with $d=3$ (and thus up to 15 nodes) can already be difficult to understand. If such \acp{DT} are used within an ensemble, the comprehensibility becomes additionally difficult. \acp{DT} with $d=1$ (decision stumps) are naturally very comprehensible, however, the limited expressiveness diminishes their usefulness. 
Therefore, \acp{DT} of $d=2$---thus, a maximum number of seven nodes---seem advisable. 
Regarding the number of experts, we assume that 2--5 experts are best suited for interpretability.
Alternatively, it is also computationally feasible to simply test multiple configurations of \ac{MoDT} with different numbers for $d$ and $e$. 

If an \ac{MoDT} model is used for prediction, potentially as a surrogate model of a more complex ML model like an artificial neural network, every decision can be retraced by a human. 
This allows experts to decide whether the prediction rules make sense. Additionally, in scenarios where potential misclassifications are dangerous, experts can rule out the possibility that the model will output certain predictions. 

\ac{MoDT} usually outperforms a \ac{DT} with comparable complexity regarding prediction accuracy. 
Still, the model can remain completely comprehensible for humans if the 2D gating function is selected. The separation into smaller subregions reduces the cognitive load for a human. However, in edge cases, the visualization of the gating function can potentially lead to inaccuracies when it is used to retrace a decision of the model, e.g., when a data point is close to the border of two regions. Although this scenario is rare, the users of \ac{MoDT} are implicitly enforced to incorporate this potential inaccuracy in their analysis. 


It can be seen as a disadvantage for interpretability that the generated insights are not necessarily valid for the complete data, but only for distinct subregions. It might instead seem more desirable to have an analysis that generalizes to the complete dataset, e.g., by training just a single \ac{DT} on the complete data. However, when the complexity of a \ac{DT} must be limited such that it can be comprehensible by humans, the resulting \ac{DT} might represent the real-world scenario poorly. In fact, a single short \ac{DT} might be misleading as it can overly generalize complex relationships. Humans are prone to accept just one or two reasons as the sole explanations of possibly complex problems \cite{Miller2019}.
Therefore, a single short \ac{DT} might lead humans to a rather incorrect analysis. Instead, with \ac{MoDT}, a user is forced to reflect on multiple explanations for multiple subproblems. Possibly, this is a more realistic scenario for real-world problems.

\ac{MoDT} has similarities to the commonly used IML method LIME \cite{lime}. Both methods can be used for explaining decisions in local subregions. Yet, MoDT also enables global explanation as it does not need to fit a separate surrogate model for each observation. Further, \ac{MoDT}'s explanations are stable, i.e., identical inputs lead to identical explanations.


Generally, \ac{MoDT} is not suited for all types of datasets and problems, respectively.
If a dataset can be represented appropriately by a single \ac{DT}, it does not make much sense to use \ac{MoDT} instead of a \ac{DT}. 
When \ac{MoDT} is used on such a problem, it usually reduces itself to a single \ac{DT}, regardless of the specified number of experts.
Conversely, if an individual \ac{DT} of arbitrary complexity seems entirely unsuited for a dataset, i.e., when a tree-based method cannot learn any meaningful pattern, \ac{MoDT} is likely also not a good choice for the dataset. \ac{MoDT} excels when a problem can be divided into multiple subproblems that incorporate different input-output relationships. Here, \ac{MoDT} can outperform regular \acp{DT} as its gating function is more powerful and more flexible. 

Although \acp{DT} are usually among the first examples of interpretable methods in IML literature, it is often not precisely defined how a \ac{DT} can enable interpretability. While there are many implementations and extensions of \acp{DT}, most of them do not consider interpretability but purely focus on performance. 
We believe that there is potential for the improvement of DT algorithms and associated visualization methods. The algorithms should not treat interpretability as a byproduct but should be designed specifically with interpretability in mind.
As \ac{MoDT} builds upon \acp{DT}, \ac{MoDT} would directly benefit from any improvements. 


While there are numerous ways to improve \ac{MoDT} from a technical perspective, we suggest that further work should initially focus on evaluating the method's benefit for interpretability in practice. This could be done with user surveys \cite{Piltaver2016} or in-depth case studies. Ideally, the studies compare \ac{MoDT} with regular \acp{DT} and a wide range of other IML methods~\cite{Molnar2019}. For this, we believe that many existing studies wrongfully reduce interpretability to single-dimensional properties like comprehensibility, trust, correctness or usefulness; but in practice, interpretability should be seen as the combination of multiple and sometimes competing objectives.


\section{Conclusion}
\label{sec:conclusion}
To the best of our knowledge, \ac{MoDT} is the first ready-to-be-used \ac{DT} ensemble method based on the \ac{MoE} architecture that is specifically tailored for interpretability. The method employs a linear gating function that divides the input into disjoint subregions that are solved by distinct \acp {DT}. A novel approach is introduced that optionally limits the complexity of the gating function and thereby facilitates interpretability. Internally, the implementation uses a variation of the \ac{EM} algorithm that employs a linear regression for finding model parameters instead of a more complex gradient-based optimization method which is typically found in \ac{MoE} architectures. Enabling interpretability, \ac{MoDT} provides a method to visualize the gating function which is adjusted to match the plots of the \acp{DT}.
The experiments show that the implementation works and consistently outperforms individual \acp{DT} with similar complexity. Notably, the especially interpretable and simplistic \ac{2D} gating function can often compete with the more complex \acl{FG} function. 


\blind{%
\section*{Acknowledgment}
This work was partially supported by the Baden-Wuerttemberg Ministry for Economic Affairs, Labour and Tourism (Project KI-Fortschrittszentrum
``Lernende Syteme und Kognitive Robotik'').
}

\balance








%
%

\onecolumn
\section*{Appendix}
In Fig.~\ref{fig:gating-uci} the 2D gating functions of \ac{MoDT} with $d=2$ and $e=3$ for all datasets listed in Table \ref{tab:datasets} is plotted.

\begin{figure}[H]
    \centering
    \subfloat[Abalone dataset.]{%
    \includegraphics[width=0.46\columnwidth]{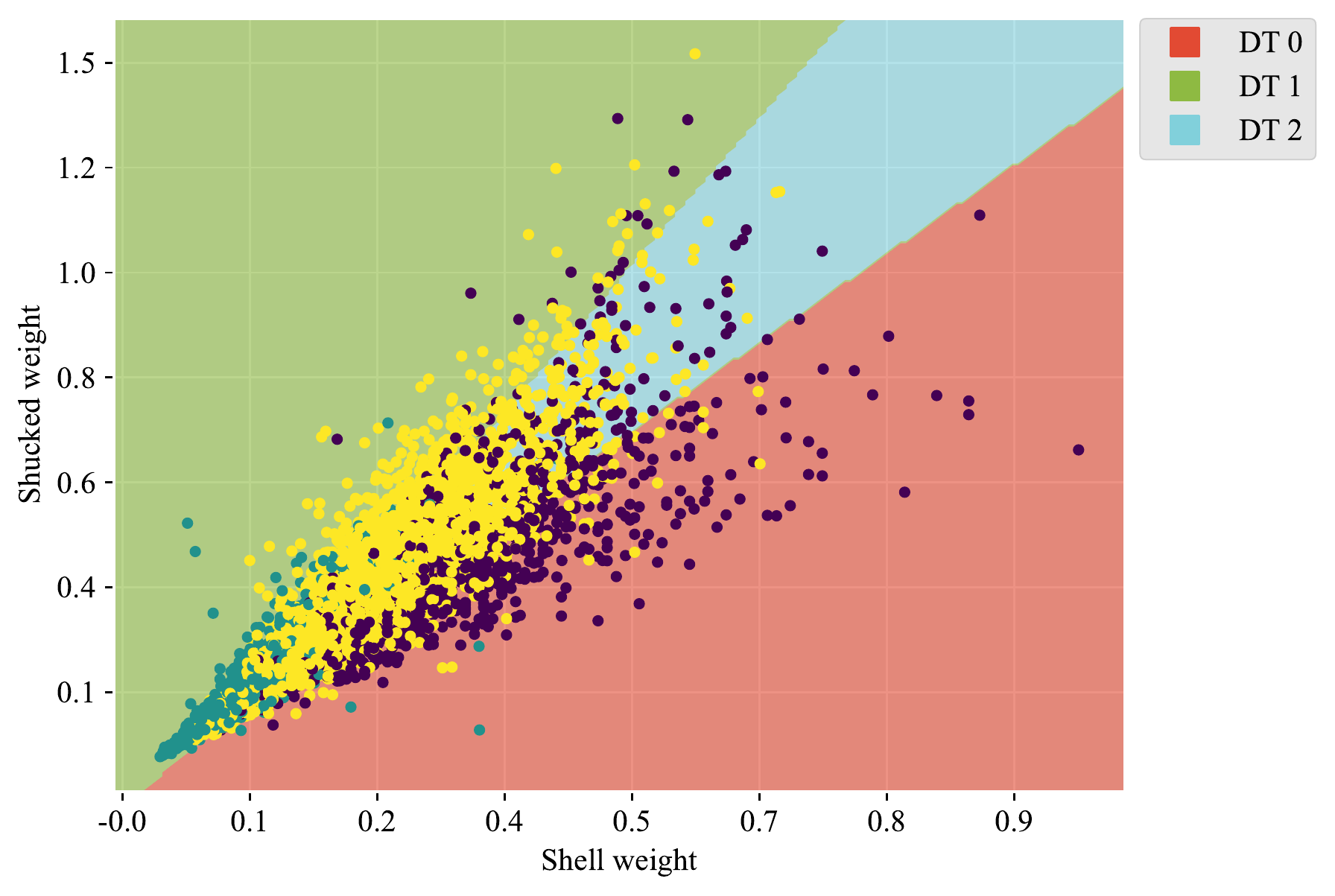}
    }\qquad%
    \subfloat[Banknote dataset.]{%
    \includegraphics[width=0.46\columnwidth]{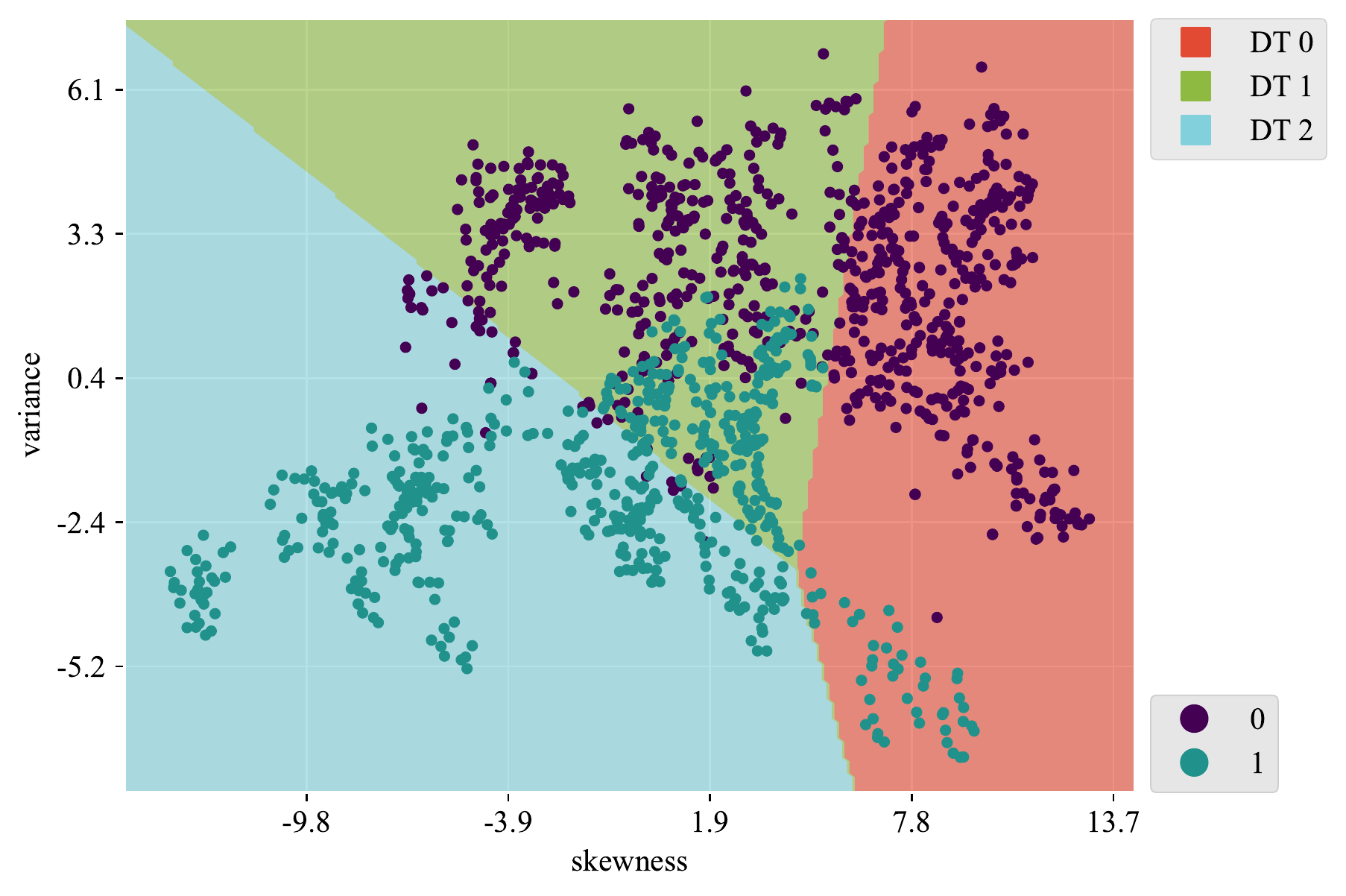}
    }\\
    \subfloat[Breat dataset.]{%
    \includegraphics[width=0.46\columnwidth]{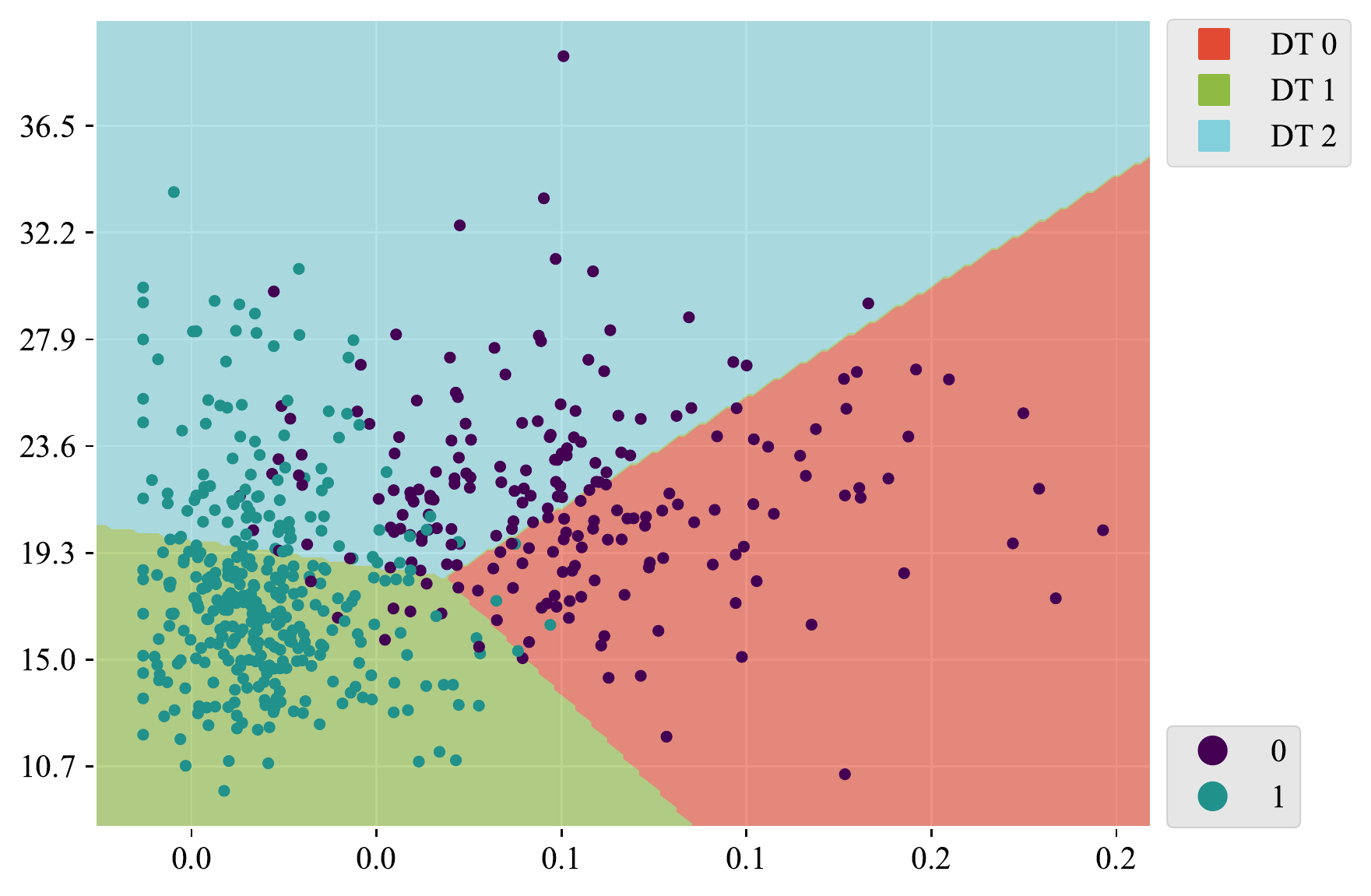}
    }\qquad%
    \subfloat[Cars dataset. It is worth mentioning that the dataset only incorporates two values for the selected gating dimensions.]{
    \includegraphics[width=0.46\columnwidth]{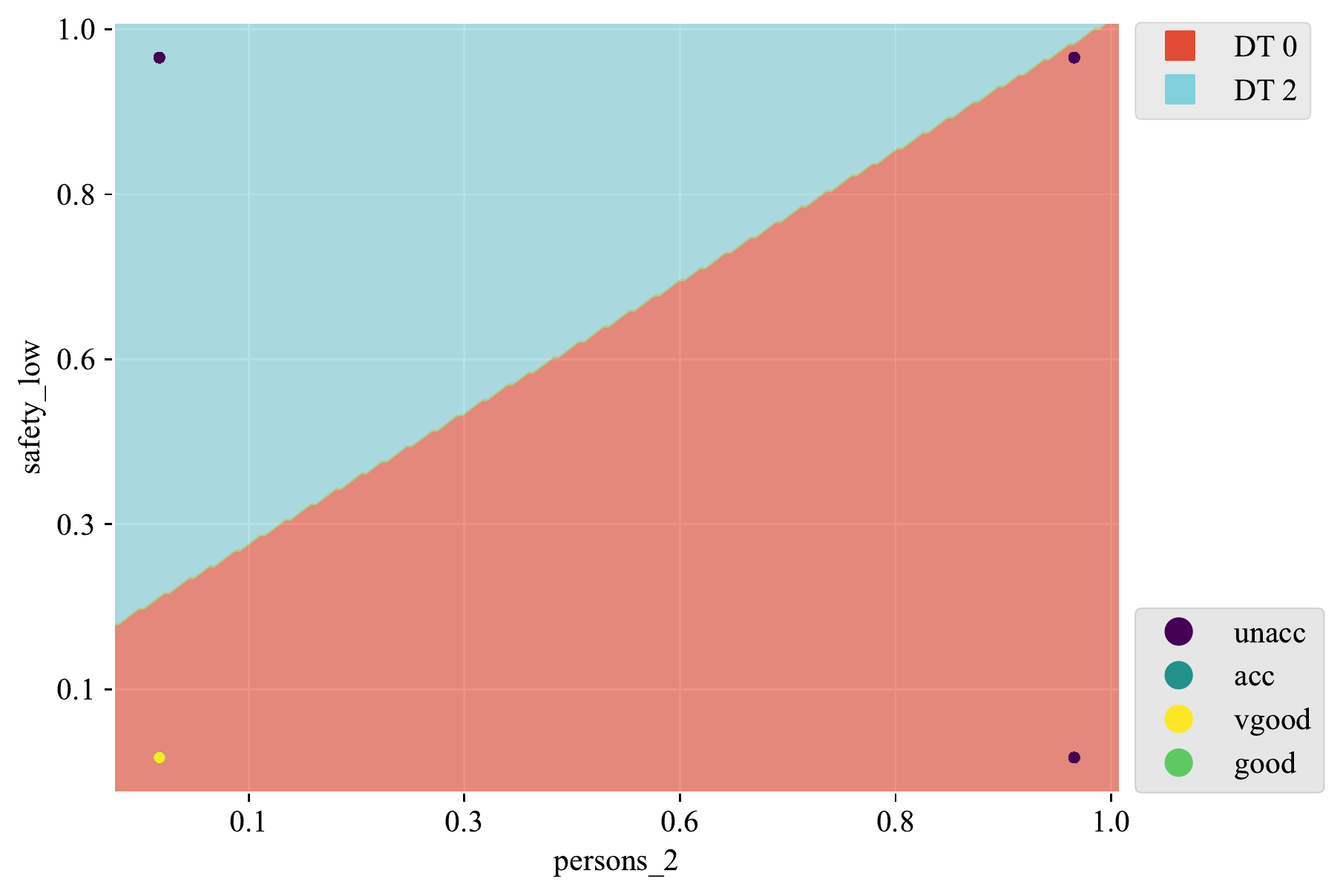}
    }\\
    \subfloat[Contraceptive dataset.]{
    \includegraphics[width=0.46\columnwidth]{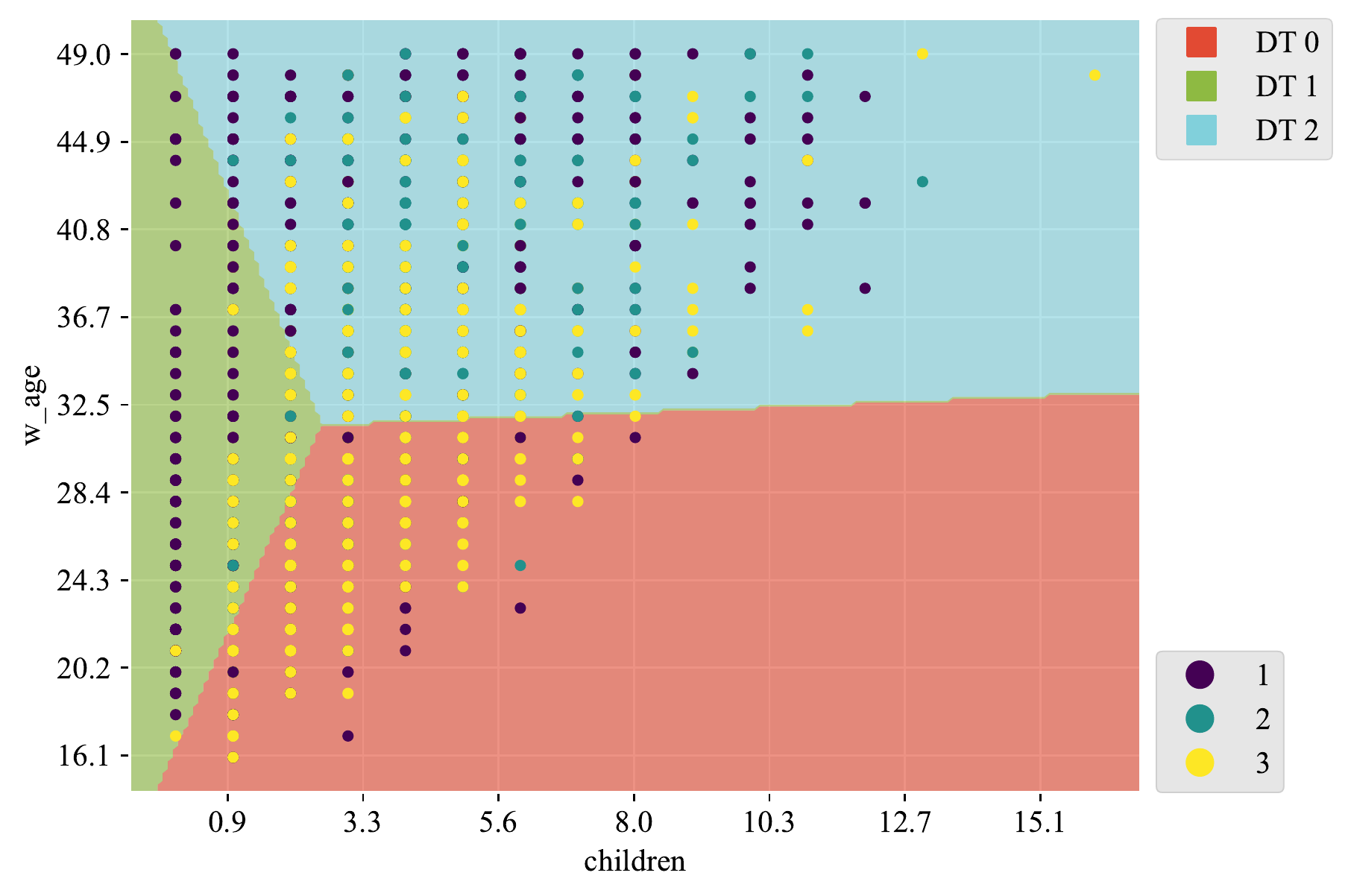}
    }\qquad%
    \subfloat[Iris dataset.]{
    \includegraphics[width=0.46\columnwidth]{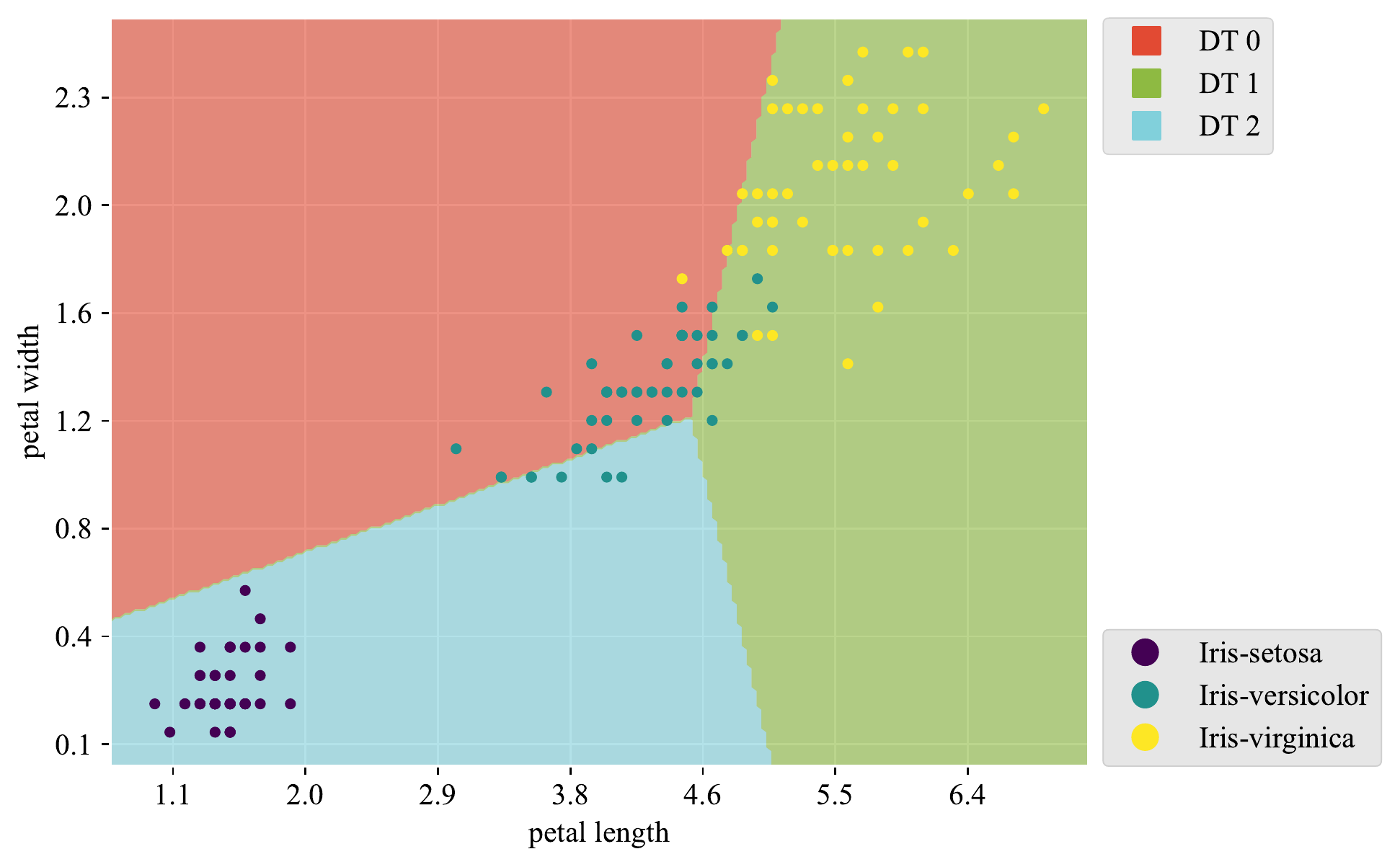}
    }
    \caption{2D gating functions of \ac{MoDT}.}
    \label{fig:gating-uci}
\end{figure}
\begin{figure}[h]
    \ContinuedFloat
    \subfloat[Steel dataset.]{
    \includegraphics[width=0.46\columnwidth]{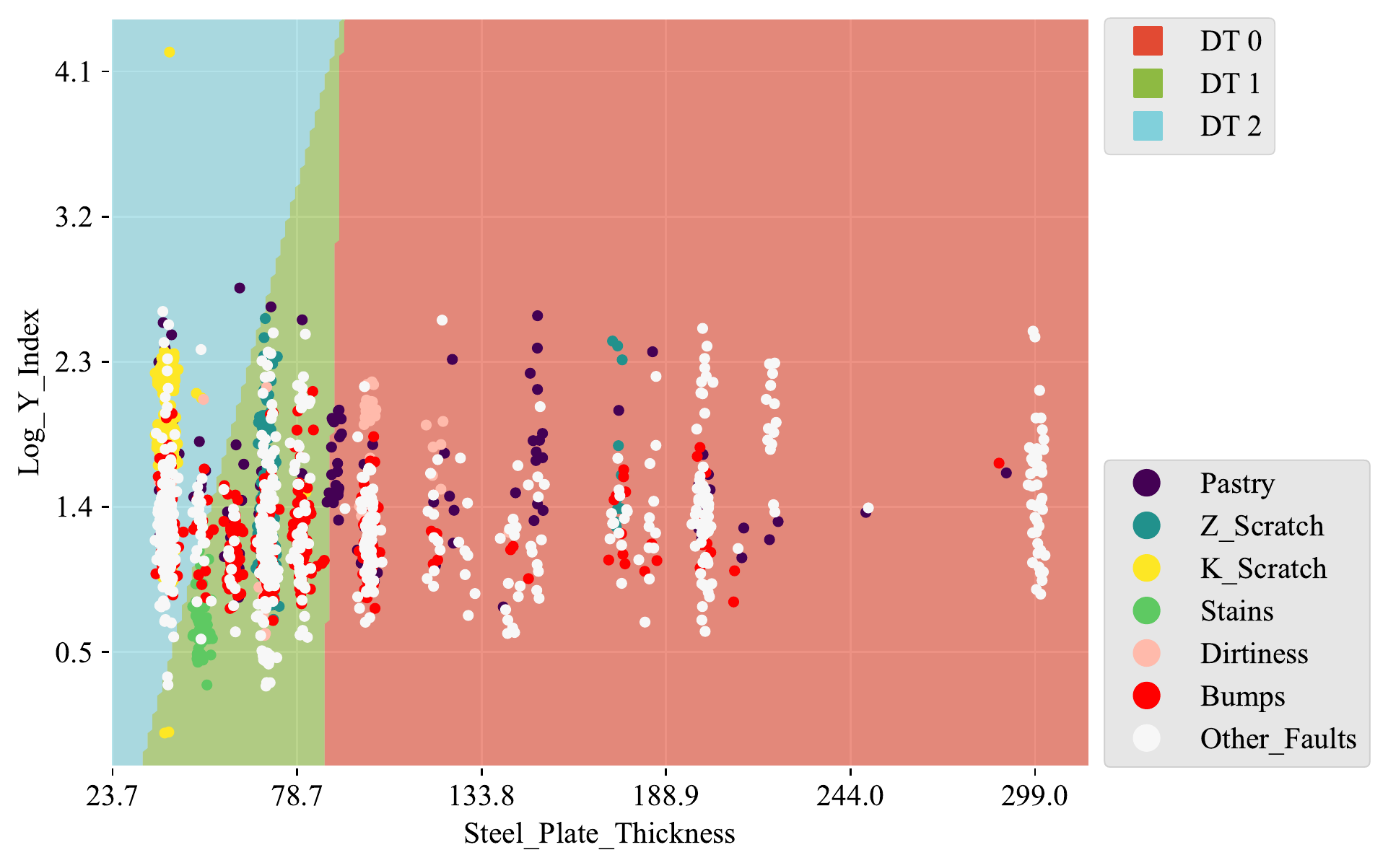}
    }\qquad%
    \subfloat[Students dataset. It is worth mentioning that the dataset only incorporates two values for the selected gating dimensions.]{
    \includegraphics[width=0.46\columnwidth]{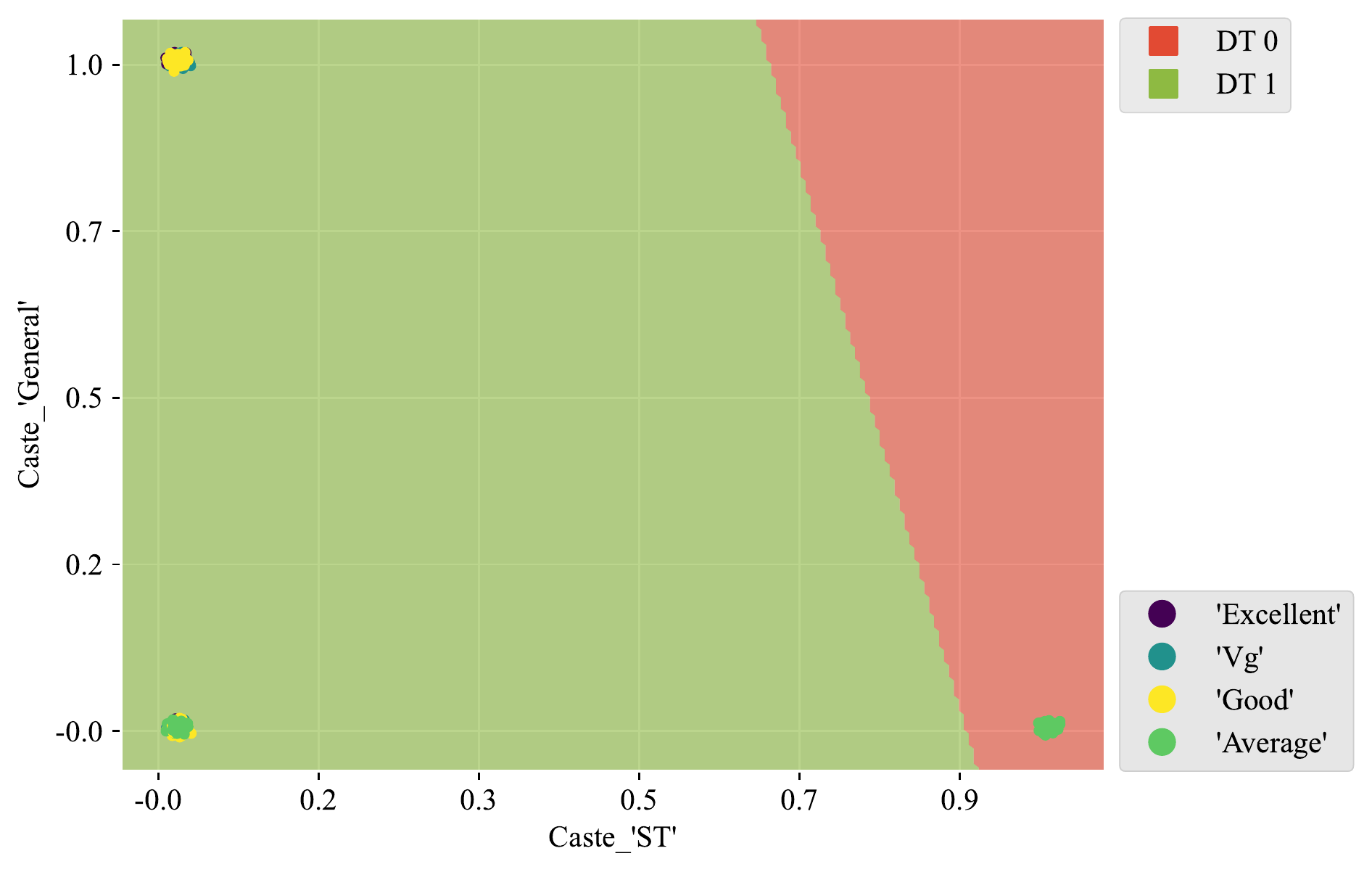}
    }
    \caption{2D gating functions of \ac{MoDT}.}
    \label{fig:gating-uci}
    \vspace*{17cm}
\end{figure}

\end{document}